\documentclass[runningheads]{llncs}

\PassOptionsToPackage{table,xcdraw}{xcolor}

\usepackage{eccv}

\usepackage{eccvabbrv}

\usepackage{graphicx}
\usepackage{booktabs}
\usepackage{colortbl}
\usepackage{adjustbox}
\usepackage{xspace}
\usepackage{multirow}
\usepackage{tabularx}
\usepackage{caption}
\usepackage{subcaption}
\usepackage{etoolbox}
\usepackage{tcolorbox}
\usepackage{amssymb}
\usepackage{titletoc}
\usepackage[accsupp]{axessibility}  

\DeclareRobustCommand{\ours}{IDeaL\xspace}
\DeclareRobustCommand{\dino}{DINO\xspace}
\DeclareRobustCommand{\deit}{DeiT-3\xspace}
\DeclareRobustCommand{\dbot}{dBOT-ft\xspace}
\DeclareRobustCommand{\ibot}{iBOT\xspace}
\DeclareRobustCommand{\imagenet}{ImageNet\xspace}
\DeclareRobustCommand{\fractals}{FractalDB\xspace}
\newcommand{\nt}{n}                                
\newcommand{\te}[1][t]{T_{#1}}                     
\newcommand{\st}{S}                                
\newcommand{\f}[1]{f^{#1}}                         
\newcommand{\ft}[1][t]{f_{\text{T}_{#1}}}          
\newcommand{\fs}{f_\text{S}}                       

\newcommand{\img}{I}                               
\newcommand{\fcls}[1]{\mathbf{f}_{#1}}             
\newcommand{\fclss}{\mathbf{f}_\text{S}}           
\newcommand{\fclst}[1]{\mathbf{f}_{\text{T}_#1}}   
\newcommand{\fpatch}{\mathbf{F}}                   
\newcommand{\fpatchs}{\mathbf{F}_{\text{S}}}       
\newcommand{\fpatcht}[1]{\mathbf{F}_{\text{T}_#1}} 
\newcommand{\np}{P}                                
\newcommand{\nps}{P_{\text{S}}}                    
\newcommand{\npt}[1]{P_{\text{T}_#1}}              
\newcommand{\fdim}[1]{d}                           
\newcommand{\fdims}{d_\text{S}}                    
\newcommand{\fdimt}[1]{d_{\text{T}_#1}}            
\newcommand{\Ldist}{\mathcal{L}^\text{Dist}}       
\newcommand{\Lte}[1]{\mathcal{L}^\text{Dist}_{#1}} 
\newcommand{\Lcos}{\mathcal{L}_{\text{cos}}}       
\newcommand{\Lsl}{\mathcal{L}_{\text{sl1}}}        
\newcommand{\LG}{\mathcal{L}^\text{G}}                    
\newcommand{\LTV}{\mathcal{L}^\text{TV}}                
\newcommand{\LPD}{\mathcal{L}^\text{PD}}                
\newcommand{\LID}{\mathcal{L}^\text{ID}}              
\newcommand{\I}{\mathbf{I}}                        
\newcommand{\R}{\mathbb{R}}                        
\NewDocumentCommand{\cossimmat}{O{i} O{\ell}}{\Gamma_{#1}^{#2}} 

\usepackage{hyperref}

\usepackage{orcidlink}

\renewcommand{\paragraph}[1]{\medskip\noindent\textbf{#1}}
\newcommand{\gain}[1]{\cellcolor[HTML]{FFE9D2}{#1}} 
\newcommand{\gaintext}[1]{\colorbox[HTML]{FFE9D2}{#1}}
\newcommand{\improv}[1]{\makebox[1.8em][l]{\textcolor{OliveGreen}{\tiny\ensuremath{\uparrow}#1\%}}}

\tcbset{colback=gray!10,colframe=black!30,boxrule=0.6pt,arc=2pt,left=2pt,right=2pt,top=2pt,bottom=2pt}
\newtcolorbox{keyobs}{}
\newcommand{\invismidrule}{%
\arrayrulecolor{white}\midrule\arrayrulecolor{black}%
}

\newcounter{datarow}
\newcommand{\datarownumber}{\stepcounter{datarow}\arabic{datarow}}

\begin{document}
\title{IDeaL: Data-Free Multi-Teacher Distillation \\
via Improved Dead Leaves}
\titlerunning{IDeaL: Data-Free Multi-Teacher Distillation}

\author{Feyza Yavuz
\and
Mert Bülent Sarıyıldız 
\and
Diane Larlus
}

\authorrunning{F.~Yavuz et al.}

\institute{NAVER LABS Europe}

\maketitle
\begin{abstract}
Multi-teacher distillation has emerged as a way to combine complementary teacher models into a single student model that exhibits the strengths of all its teachers.
The student is trained to mimic the output of the teachers on a set of images, typically the union of the individual teacher's training sets, assuming this data is available.
In this paper, we question that assumption and explore alternative options.
We first study how far one can go when distilling from teachers fed with different types of noise.
Then, we show that information contained in the teachers can be leveraged to tailor the noise for multi-teacher distillation: we propose a method that, thanks to decorrelation losses at both patch and image levels, generates teacher-specific, improved samples optimized for data-free distillation.
Experiments show that our most effective samples, \ours, lead to strong students that successfully capture complementary information from the teachers, yielding surprisingly competitive results 
that substantially narrow the gap with
students distilled from real images.
Moreover, given a limited budget of 1K images for distillation, students distilled using our \ours samples match or surpass the performance of those distilled using a 1K-image subset of \imagenet.
\end{abstract}
\section{Introduction}
\label{sec:intro}

Vision Transformers (ViTs)~\cite{dosovitskiy2021an} pretrained on large collections of web images have become the standard visual backbone, largely replacing CNNs.
Today, it is common practice to use such pretrained visual encoders, which provide general-purpose feature representations to a wide range of downstream tasks.
However, different models capture distinct visual characteristics, depending on their architecture, training process, or training data.
Multi-teacher distillation has thus emerged as a popular way to reconcile these complementary views~\cite{ranzinger2024amradio,roth2024fantastic,sariyildiz2025dune,sariyildiz2024unic,shi2023hybrid,tian2020contrastive,ypsilantis2024udon,heinrich25radiov2.5}; multiple teacher models jointly guide the learning of a single student, allowing it to inherit knowledge from all teachers at once. 

However, multi-teacher distillation methods typically require large amounts of real images during training.
These methods try to cover the input domain of the teachers by using their training data when available~\cite{roth2024fantastic,sariyildiz2024unic,sariyildiz2025dune} or by using a large web-based dataset such as DataComp-1B~\cite{gadre2023search} as in~\cite{ranzinger2024amradio,heinrich25radiov2.5}.
This reliance on real data poses practical challenges when the teachers' original training data is unavailable or cannot be redistributed due to legal restrictions, privacy concerns, or proprietary data policies.
This issue is becoming relevant as foundation models are increasingly trained on proprietary data that is not publicly accessible \cite{simeoni2025dinov3,tschannen2025siglip}.
As a result, obtaining suitable real-image datasets for distillation can be costly, impractical, or simply not possible.

\begin{figure}[t!]
    \centering
    \includegraphics[width=\columnwidth]{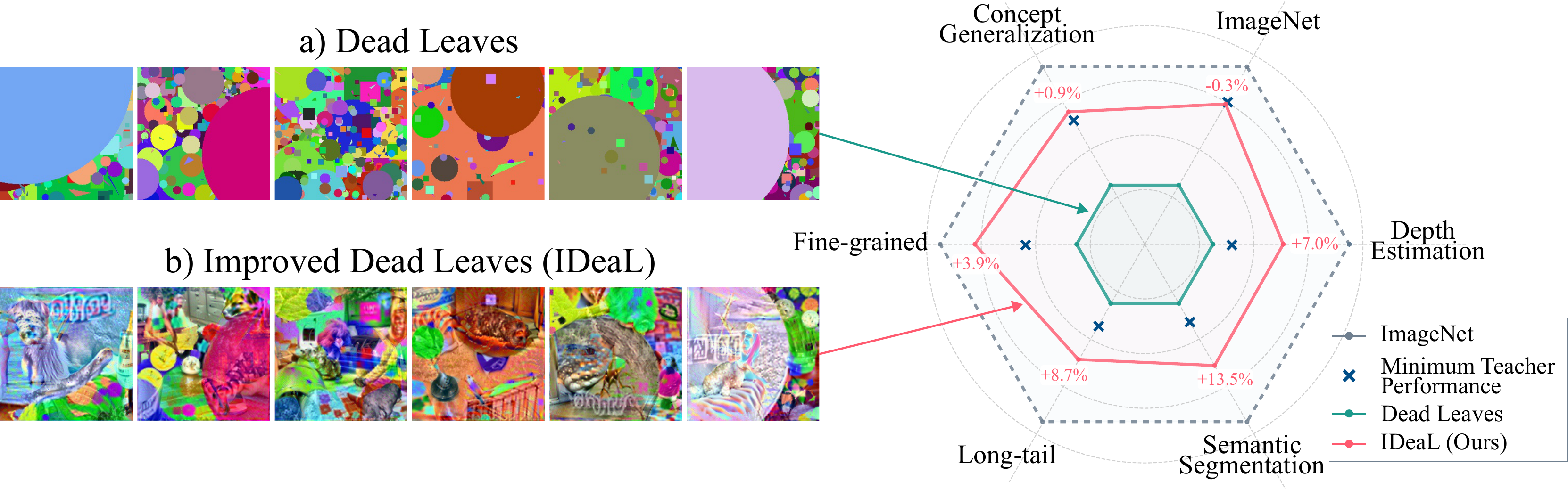}
    \caption{
        In this paper, we distill complementary teachers into strong models, even without real data.
        \textbf{(Left)} Examples of the noisy samples that we use for distillation: i) procedurally generated synthetic Dead Leaves and ii) its improved version \ours. 
        \textbf{(Right)} Performance on several downstream tasks of a student distilled  from 4 teachers using 1M samples of either \imagenet, Dead Leaves, or our Improved Dead Leaves (\ours). 
        We report the minimum performance obtained by the teachers on each task as a reference 
        and denote the relative gains obtained by using \ours compared to the minimum teacher.
    }
    \label{fig:teaser}
\end{figure}

In this paper we take an adversarial stance and ask:
{\emph{how far can multi-teacher distillation go if not provided with any data?}}
Earlier work~\cite{sariyildiz2024unic} has taken one step in this direction and replaced
\imagenet \cite{russakovsky2015ilsvrc}
with {realistic}
synthetic variants {for each \imagenet class} generated by Stable Diffusion~\cite{sariyildiz2023fake}.
We go further and pursue a completely {\em data-free} approach, where images are entirely replaced by procedurally-generated, synthetic surrogates.
We start by analyzing how distillation performance degrades as the number of \imagenet samples decreases.
We then replace real images with structured noise generated with procedural programs and compare several options already considered in data-free pretraining~\cite{kataoka20pre-training, nakashima21can, baradad21learning}. 
Among these, we find that Dead Leaves~\cite{baradad21learning} consistently outperforms other alternatives, and shows encouraging transfer performance.

Building on these promising results, we propose a pixel-level optimization method that injects information from teacher attention layers into synthetic images, tailoring them for multi-teacher distillation.
Starting from the best data-free alternative, Dead Leaves, we optimize these samples so that patches exhibit diverse representations within a single synthetic image while global representations differ across these images.
To achieve this, we introduce decorrelation losses that encourage low cosine similarity between patches within an image and between image-level representations, yielding richer and more diverse inputs for distillation.
Note that our synthetic images are not meant to recover the teachers' training sets, not even to look realistic.
Instead, they are optimized to capture how the teachers process the data.
This allows the student to learn {from} the teachers' complementary views, even without  data.

To properly evaluate the effectiveness of our generation method and the resulting synthetic samples, we need a controlled distillation and evaluation setting.
We choose to follow UNIC~\cite{sariyildiz2024unic} as our main framework for multi-teacher distillation.
In this framework, \imagenet is the dataset used by all teachers for training and distillation.
That way, when we systematically replace it with synthetic alternatives, we can isolate the impact {of our design choices} on student performance.
Moreover, the framework evaluates on a diverse set of classification and dense prediction tasks, providing a comprehensive overview of the distilled students' performance.

To summarize, our contributions are as follows.
First, we study data-free multi-teacher distillation, systematically analyzing how distillation performance varies when reducing the amount of real data, and replacing it with procedurally generated synthetic datasets.
Second, we propose a pixel-level optimization method, tailored to ViTs \cite{dosovitskiy2021an}, that enriches synthetic images using complementary information provided by multiple teachers.
Applied to the synthetic Dead Leaves samples, it produces a much more effective training set of Improved Dead Leaves (or \ours in short) for multi-teacher distillation.

Through extensive experiments in a controlled setting, we show that \ours samples yield strong students that close much of the performance gap with the students distilled on real images.
Given a budget of only 1K images, they even match or surpass them.
With 1M samples, they lead to strong students that are versatile enough to be competitive on all tasks, outperforming the minimum teacher performance on all considered tasks (except for \imagenet, the training set of the teachers), as shown in \cref{fig:teaser}.
To the best of our knowledge, this is the first work that fully replaces real images with synthetic surrogates in multi-teacher distillation while maintaining competitive performance across a large number of downstream tasks.
\section{Related Work}\label{sec:relwork}

In this section, we briefly review the literature relevant to our work. We first discuss recent knowledge distillation approaches and how those relate to our pipeline.
Then we provide a short overview of methods that generate purely synthetic data tailored for different tasks or models.

\paragraph{Knowledge Distillation (KD)} 
was introduced as a compression mechanism~\cite{bucilua2006model} to train a typically smaller student model using the outputs or intermediate features from a larger teacher model~\cite{romero2015fitnets,zagoruyko2017paying,hinton2014distilling}.
{\em Multi-teacher} distillation~\cite{ranzinger2024amradio,roth2024fantastic,sariyildiz2025dune,sariyildiz2024unic,shi2023hybrid,tian2020contrastive,ypsilantis2024udon, heinrich25radiov2.5} extends this idea by training a student to align with multiple teachers, each potentially pretrained under different objectives or data distributions, allowing it to build a unified representation suitable for many tasks.
For effective distillation, prior work uses the same data for training the teachers and distilling to the student~\cite{roth2024fantastic,sariyildiz2024unic,sariyildiz2025dune}. When teacher data is unavailable,~\cite{ranzinger2024amradio, heinrich25radiov2.5} resort to large web-based datasets~\cite{gadre2023search} to cover the teachers' input domain.
In both cases, these methods rely on large collections of real images.
In contrast, our approach performs multi-teacher distillation \emph{without any real images}: we use only procedurally generated synthetic data, optimized via teacher attention features to be maximally informative for all teachers simultaneously.

\paragraph{Data-free knowledge distillation} was introduced~\cite{lopes2017datafree, nayak2019zero, micaelli2019zero} to tackle the problem of teacher data not being available during distillation.
A central idea is to exploit the teacher's knowledge to synthesize data that can be used for distillation, which often comes with strong assumptions about \eg the availability of teacher's training data or class label information.
For instance,~\cite{lopes2017datafree}~keeps a record of the teacher's activations extracted on the original training data. \cite{chen2019dafl,micaelli2019zero,nayak2019zero,ye2020datafree,fang2020datafree,patel2023learning} rely on class logits predicted by the teacher to train a generator that produces synthetic data, assuming the teacher was trained on a classification task, potentially requiring balanced data per class~\cite{chen2019dafl}.
We avoid such assumptions, as we require neither teachers to be classification models nor any label information.
Moreover, our approach does not suffer from the potential instability issues due to co-training a generator along with the student with an adversarial objective~\cite{patel2023learning}.

Other works decode the knowledge embedded in the teacher's internal statistics, such as BatchNorm (BN)\cite{ioffe2015batchnormalization}, to guide data synthesis.
In this context, \cite{yin2020dreaming,fang2021contrastive,zhange2021diversifying,yin2021see,hu2024sparse} synthesize class-conditioned images by matching BatchNorm statistics stored in a pretrained CNN.
\cite{raikwar22discovering} mitigates the activation covariate shift when providing Gaussian noise for distillation by keeping the teacher's BN modules in ``training'' mode.
However, real data from the teacher's input domain is then needed at inference time to compute ``running statistics'' for the student's BN layers. Given that ViTs~\cite{dosovitskiy2021an} use LayerNorm~\cite{ba2016layer} instead of BN, it is not straightforward to apply these methods to modern ViT-based teachers.

Another line of work focuses on data-free model quantization of ViTs.
PSAQ-ViT~\cite{li2022patch} is among the first works that propose a patch-similarity-based regularization for ViT-based models. It optimizes images from Gaussian noise, and subsequent works improve the inversion quality or efficiency~\cite{li2024toward, hu2024sparse, zhong2025semantics, ramachandran2024clampvit, zhao2025enhancing, choi2025mimiq}.
However, these ViT-targeted inversion methods are designed for a \emph{single} model and for a different goal (\eg for quantization) which requires only small-scale sample sets, making them difficult to generalize to multi-teacher settings where the student must learn from diverse teachers simultaneously.
Our method differs from these works as it is designed from the ground up for the \emph{multi-teacher} setting, jointly optimizing large-scale synthetic data across all teachers.

\paragraph{Procedural synthetic data.}
Recent work has explored formula-driven, procedurally generated data as alternatives to real images for pretraining~\cite{kataoka20pre-training, nakashima21can, baradad21learning, baradad2022procedural, johnson2017clevr, zhai2020largescale, guss2019minerl}, domain adaptation~\cite{tang2023anew}, and knowledge distillation~\cite{frank2023data, frank25what}.
The premise is that such datasets enable truly data-free learning. They are not the output of generative models~\cite{fan2024scaling, sariyildiz2023fake} trained to produce realistic-looking synthetic images from real ones.
\fractals~\cite{kataoka20pre-training} was introduced as formula-driven supervised learning and quickly extended~\cite{nakashima21can, kataoka22replacing, nakamura23pre-training, nakamura24scaling, ryosuke24formula, kataoka21formula, yamada21MV-FractalDB, takashima23visual} to the point where fractals can replace real images for pretraining~\cite{nakamura23pre-training}.
In parallel, Baradad~\etal~\cite{baradad21learning} showed that structured noise
can yield competitive representations. One example is Dead Leaves, an image model
generated using simple formulas. However, performance of models pretrained on such synthetic datasets does not scale with the data size.
Baradad~\etal~\cite{baradad2022procedural} later collected diverse procedural generation programs with better scaling properties, inspiring Frank~\etal~\cite{frank2023data} to propose a data-free KD framework using procedurally rendered images.
In follow-up work, Frank~\etal~\cite{frank25what} analyze what properties make surrogate datasets effective for KD.
All these works use procedurally generated data \emph{as-is} and focus on pretraining or single-teacher distillation.
Our approach differs in two ways: first, we study the multi-teacher setting, where the student must simultaneously learn from teachers with heterogeneous training objectives; second, rather than using procedural images directly, we take Dead Leaves as a starting point and perform direct pixel-space optimization guided by multiple teachers' attention features, tailoring the synthetic data to the specific set of teachers.

\section{Background}
\label{sec:background}

In this work, we train student models distilling from multiple pretrained teachers, where all models are ViTs~\cite{dosovitskiy2021an}. First, we
briefly introduce the ViT architecture and the multi-teacher distillation framework, and then discuss the data used for distillation, which is the main focus of this paper.

\paragraph{Vision transformers (ViTs)} process input images by dividing them into a grid of $\np$ non-overlapping patches, which are then linearly projected into a sequence of patch tokens.
A special \texttt{CLS} token is {included} in this sequence and serves as a global representation.
Multiple layers of self-attention and feed-forward blocks are then applied to these tokens, allowing the model to learn complex relationships between different parts of the image.
We denote the mapping from an input image $\img \in \R^{H \times W \times 3}$ to the output features of a ViT encoder as $\f{}(\img)$.
It produces a set of $\np$ patch features $\fpatch{} \in \R^{\np \times \fdim{}}$ and the \texttt{CLS} feature $\fcls{} \in \R^{\fdim{}}$, where $\fdim{}$ is the feature dimension of the model.

\paragraph{Multi-teacher distillation} aims at distilling a given set of $\nt$ pretrained teacher models $\{\te[1], \te[2], \ldots, \te[\nt]\}$ into a single student model $\st$.
Each teacher $\te[i]$ is a ViT encoder whose mapping $\ft[i]$ produces:
 patch features $\fpatcht{i} \in \R^{\npt{i} \times \fdimt{i}}$ and  \texttt{CLS} feature $\fclst{i} \in \R^{\fdimt{i}}$,
where $\npt{i}$ and $\fdimt{i}$ are the number of patches and feature dimension of the $i$-th teacher, respectively.
Similarly, the student model $\st$ is a ViT encoder with mapping $\fs$ that produces patch features $\fpatchs \in \R^{\nps \times \fdims}$ and a \texttt{CLS} feature $\fclss \in \R^{\fdims}$.
For the sake of simplicity, we assume that the student and all teachers have the same number of patches $\nps = \npt{i}$, and feature dimensions $\fdims = \fdimt{i}$ for all $i$.
Some teachers may not have a \texttt{CLS} token; we replace it with the global average-pooled patch features, following~\cite{sariyildiz2024unic}.
The student is trained such that its features ($\fclss$ and $\fpatchs$) match every teacher's representation ($\fclst{i}$ and $\fpatcht{i}$).
This is done by minimizing the distillation loss $\Ldist = \sum_{i=1}^{\nt} \Lte{i}$,
which sums over teacher-specific distillation losses broadly defined as $\Lte{i} = \rho^\text{CLS} (\fclss, \fclst{i}) + \rho^\text{Patch} (\fpatchs, \fpatcht{i})$, where $\rho^\text{CLS}$ and $\rho^\text{Patch}$ are similarity measures between the student and teacher features, such as cosine similarity or smooth-$\ell_1$ distance operating on the respective feature vectors.
By optimizing on this combined objective over all teachers simultaneously, the student learns a unified representation that captures complementary knowledge encoded by the teachers.
We closely follow the distillation framework from UNIC~\cite{sariyildiz2024unic}, see supplementary material for further details.

\paragraph{Distillation data.}
The loss $\Ldist$ is computed over a set of training images.
Existing multi-teacher distillation methods typically rely on large collections of real images to cover the input domain of the teachers. For instance, UNIC~\cite{sariyildiz2024unic} uses the full \imagenet training set (1.28M images) while RADIO~\cite{ranzinger2024amradio,heinrich25radiov2.5} scales to 1B images and DUNE~\cite{sariyildiz2025dune} combines more than 10 heterogeneous datasets.
However, as mentioned in the introduction, the teachers' original training data may be unavailable due to licensing, privacy, or proprietary restrictions, so in this paper we study a drastically different approach where we assume access to no real data for distillation. We explore replacing real images entirely with Gaussian noise and procedurally generated synthetic data (such as \fractals~\cite{kataoka20pre-training}, and Dead Leaves~\cite{baradad21learning}) and find that Dead Leaves consistently outperforms the other synthetic alternatives across all downstream tasks. Still, a gap remains between Dead Leaves and real data.
In the next section, we propose a method to reduce this gap by \emph{optimizing} synthetic images to be directly maximally informative of teachers, and hence better suited for multi-teacher distillation.
\begin{figure}[t]
    \centering
    \includegraphics[width=\columnwidth]{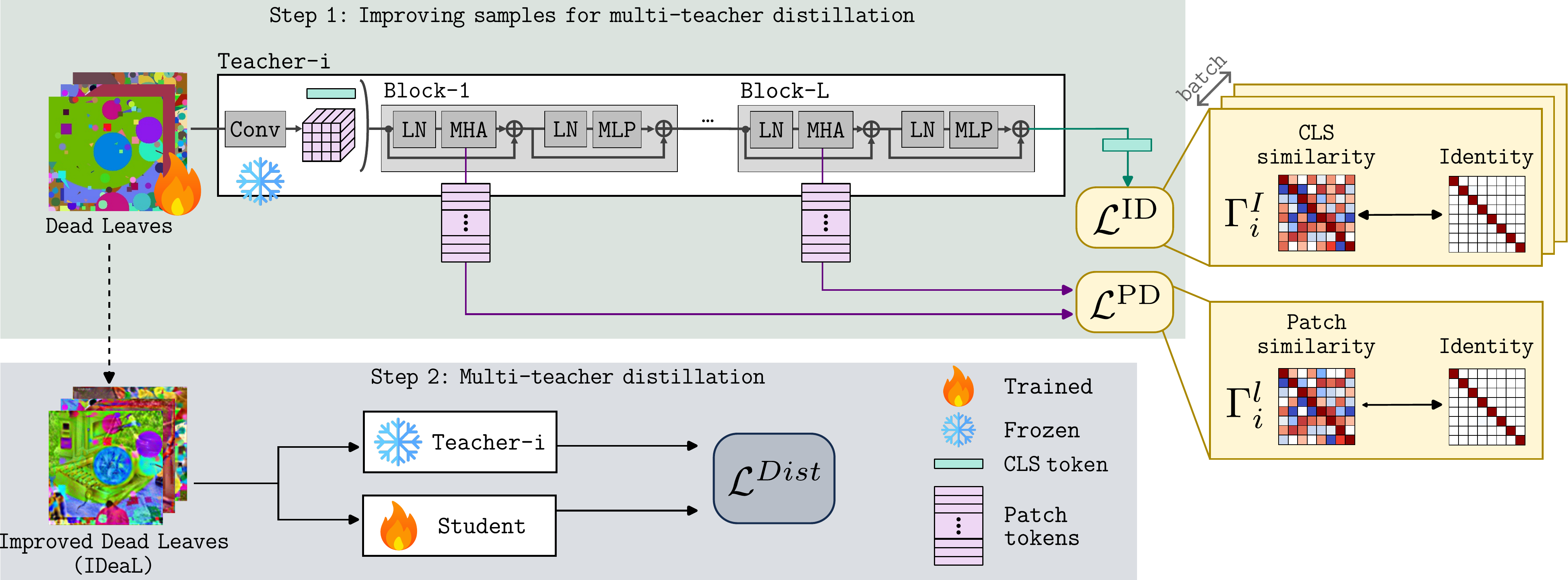}
    \caption{
        Illustration of our 2-stage method for multi-teacher distillation without real data.
        {\bf (Step 1)} Our method for improving input data tailored for multi-teacher distillation.
        Note that gradients of $\LID$ and $\LPD$ are backpropagated through the frozen teacher encoders, which allows the synthetic images to be optimized to be maximally informative for all teachers simultaneously.
        {\bf (Step 2)} Improved data is then used to train the student model distilling from all teachers, exactly as if it were real data.
    }
    \label{fig:method_figure}
\end{figure}

\section{Optimizing images for multi-teacher distillation}
\label{sec:method}

Our goal is to train student $\st$ by distilling from multiple pretrained teachers $\{\te[1],\ldots,\te[\nt]\}$ without relying on real data.
To this end, we propose a method that generates synthetic images initialized from structured noise. These images are optimized to be maximally informative for all teachers jointly, so that distilling with such data yields a strong student.
An observation from \cite{li2022patch} is that ViT self-attention layers rely on rich patch representations to capture meaningful spatial and semantic relationships within an image. However, when all patches look alike, as in Gaussian noise, the attention mechanism has little to work with.
We therefore design generation objectives that encourage both \emph{within-image} diversity, so that every patch carries distinct information, and \emph{across-image} diversity, so that different generated samples cover complementary regions of teachers' feature spaces.
These are defined as decorrelation losses operating on teacher attention features, explained in detail below.

\paragraph{Image optimization.}
Given an initial image $\img \in \R^{H \times W \times 3}$, \eg  a sample from Dead Leaves, we treat its pixel values as learnable parameters.
We then iteratively update these pixels by minimizing a generation loss $\LG$ (defined below), backpropagating through the frozen teacher encoders $\ft[1], \ldots, \ft[\nt]$.
Once optimized, the resulting synthetic images are stored and used as training data for distillation, \ie to compute the distillation loss $\Ldist$ described in \cref{sec:background}.

\paragraph{Patch decorrelation loss.}
Real images are quite complex, which leads to diverse patch representations in ViT layers.
In contrast, unstructured noise yields nearly homogeneous patch representations, limiting the expressiveness of self-attention layers.
To generate images with diverse internal structure, we encourage patch representations to be decorrelated across all teachers simultaneously.

For a given teacher $\te[i]$, we extract the attention output at each transformer layer $\ell \in \{1,\ldots,L\}$ and average across heads to obtain per-patch representations.
We then compute the pairwise cosine similarity matrix $\cossimmat[i] \in \mathbb{R}^{\np \times \np}$, where entry $(j,k)$ is the cosine similarity between the representations of the $j$-th and $k$-th patches.
Inspired by self-supervised decorrelation objectives~\cite{zbontar2021barlow,zhang2025rsd}, we define the patch decorrelation loss as:
\begin{equation}\label{eq:patch-decorr-loss}
    \LPD_i = \frac{1}{\np^2}\sum_{\ell=1}^{L} \left\lVert \cossimmat[i] - \I_\np \right\rVert_F^2,
\end{equation}
where $\I_\np \in \mathbb{R}^{\np \times \np}$ is the identity matrix and $\lVert\cdot\rVert_F$ is the Frobenius norm.
Since the cosine similarity of a patch with itself is always one, the diagonal entries of $\cossimmat[i]$ are already at the target; minimizing \cref{eq:patch-decorr-loss} therefore drives all off-diagonal similarities to zero, making patches as dissimilar as possible within each image.

\paragraph{Image decorrelation loss.}
While $\LPD$ encourages within-image diversity, it does not prevent mode collapse, where different generated samples are highly similar to each other.
To promote diversity across samples, we introduce a decorrelation loss on the global image representations.
For each teacher $\te[i]$ and for each image in the batch, we extract the final-layer \texttt{CLS} embedding (or spatially average-pooled patch embeddings) and compute the pairwise cosine similarity matrix on image level $\cossimmat[i][\text{I}] \in \mathbb{R}^{B \times B}$, where $B$ is the batch size.
Then teacher-specific image decorrelation loss becomes:
\begin{equation}\label{eq:img-decorr-loss}
    \LID_i = \frac{1}{B^2} \left\lVert \cossimmat[i][\text{I}] - \I_B \right\rVert_F^2,
\end{equation}
where $\I_B \in \mathbb{R}^{B \times B}$.
Minimizing $\LID_i$ ensures that each generated image has a distinct global representation as perceived by teacher $\te[i]$.
If a teacher does not use a learned \texttt{CLS} token, we use the global patch feature average as a surrogate, following the convention introduced in \cref{sec:background}.

\paragraph{Total image generation loss.}
We combine both decorrelation losses across all teachers and add a {regularization term $\LTV$, whose role is to {encourage smoothness by penalizing the total variation of pixel values \cite{yin2020dreaming}}}, leading to:
\begin{equation}\label{eq:generation-loss}
    \LG = \alpha_1 \LTV + \sum_{i=1}^{\nt} \bigl(\alpha_2\LPD_i + \alpha_3\,\LID_i\bigr),
\end{equation}
where $\alpha_1, \alpha_2, \alpha_3$ are {hyperparameters}.
By summing over all $\nt$ teachers, generated images are optimized to be jointly informative across all teachers' feature spaces, which is essential to achieve strong multi-teacher distillation performance.
Unlike~\cite{yin2020dreaming,li2022patch}, our method does not assume any label or access to classification heads, making the method applicable to any combination of teachers, including self-supervised ones.
The formulation relies only on matrix multiplications and Frobenius norms, which keeps backpropagation efficient and allows scaling to large synthetic datasets (up to a million samples in our experiments) across multiple teachers.

\paragraph{Training the student.}
The image generation and distillation stages are decoupled (\cref{fig:method_figure}): we first generate the full synthetic dataset by optimizing $\LG$, and then train the student by minimizing $\Ldist$ on the resulting images, exactly as if they were real data. 
This two-stage design means that the generation cost is paid once, after which the synthetic dataset can be reused across different student architectures or distillation configurations.

\section{Results}
\label{sec:exps}

\subsection{Experimental protocol}
\label{sec:exp_protocol}

Our framework consists of two main stages: synthetic sample generation and multi-teacher distillation. In the first stage, images are generated using our pixel-optimization loss, leveraging attention features from teachers to produce informative samples (\cref{sec:method}). In the second stage, these synthetic samples are used to train the student model via multi-teacher distillation (\cref{sec:background}).

We adopt the UNIC \cite{sariyildiz2024unic} multi-teacher distillation framework and evaluation protocol,
with one critical modification: we replace the \imagenet dataset with different sets of synthetically generated samples.

\paragraph{Teachers and student.} 
Following~\cite{sariyildiz2024unic}, we distill from four ViT-B/16 teachers for 100 epochs, each pretrained on \imagenet \cite{deng2009imagenet}: two self-supervised models, (i)~\dino \cite{caron2021emerging} and (ii)~\ibot \cite{zhou2022ibot}, and two supervised models, (iii)~\deit \cite{touvron2022deitiii} and (iv)~a fine-tuned dBOT \cite{liu2024dbot} optimized for \imagenet classification, namely \dbot.
All teachers use 768-dimensional features, 12 blocks, 224$\times$224 input size and patch size 16. Importantly, all teachers are trained only on \imagenet \cite{deng2009imagenet}, which ensures a controlled experimental setup, eliminating from our analysis any bias that could come from the fact that teachers have been trained on different data. For instance, differences that we will see in generated synthetic images (\cref{fig:ablation_loss}) can be truly attributed to differences in the flavors of our method.
The student is a ViT-B/16 model with the same architecture as the teachers.

\paragraph{Sample generation.} We initialize the pixel optimization process with samples from Dead Leaves \cite{baradad21learning},
resized to $224 \times 224$.
All teachers are kept frozen, and only image pixels are trained. We use pools of 250 samples and optimize on randomly selected subsets of 40 samples. Each subset is used for 10 iterations before resampling, and each pool is optimized for a total of 4000 iterations. This increases interactions among examples within the samples while remaining computationally efficient.
At each iteration, we apply random horizontal flips \cite{yin2020dreaming,li2022patch} to improve robustness. Teacher features are normalized at the end of the forward pass. 
$\LPD$, $\LID$, and $\LTV$ are weighted by 1, 1, and 0.05, respectively.
We use the Adam \cite{kingma2017adam} optimizer, with a learning rate set to 0.1. Unless stated otherwise, we generate 1K, 10K, 100K, and 1M samples for the main experiments (\cref{tab:all_results}) and 10K samples per configuration for ablations.

\paragraph{Dataset for distillation.} We compare the performance of student models distilled from the same teachers, but using one of these five different data sources: \imagenet~\cite{russakovsky2015ilsvrc}, Gaussian noise, \fractals~\cite{kataoka20pre-training}, Dead Leaves~\cite{baradad21learning}, and improved Dead Leaves (\ours).
For each source, we consider subsets of size 1K, 10K, 100K, and 1M samples, drawn uniformly at random, except for \imagenet, where we enforce a minimum of one sample per class.
For each subset size, we report the average performance over three subsets obtained by different seeds.

\begin{table}[ht!]
\centering
\caption{
Performance of teacher models and different students distilled from them, across different distillation data sources and sizes.
We report Top-1 accuracy on \imagenet, average Top-1 accuracy on 15 classification tasks for transfer,  mIoU on ADE20K for segmentation, and RMSE on NYUd for depth estimation.
Best teacher performance and best student performance per distillation set size in \textbf{bold}.
Student performance better than the min teacher is \gaintext{highlighted in beige}.
Relative improvements of \ours over Dead Leaves are shown in green. For students distilled on \imagenet, Dead Leaves and \ours subsets, we report mean over 3 different subsets.}

\setcounter{datarow}{0}
\setlength{\tabcolsep}{2pt}
\adjustbox{max width=\textwidth}{

\begin{tabular}{r l l r p{0.25cm}@{}c@{}c p{0.25cm}@{}c@{}c p{0.25cm}@{}c@{}c p{0.25cm}@{}c@{}c}
\toprule
& \multirow{2}{1.5cm}{Model} & \multicolumn{2}{c}{Distillation data} & \multicolumn{3}{c}{\imagenet} & \multicolumn{3}{c}{Transfer} & \multicolumn{3}{c}{Seg.} & \multicolumn{3}{c}{Depth} \\
& & Source & Size & \multicolumn{3}{c}{Top-1($\uparrow$)} & \multicolumn{3}{c}{Top-1($\uparrow$)} & \multicolumn{3}{c}{mIoU($\uparrow$)} & \multicolumn{3}{c}{RMSE($\downarrow$)} \\
\midrule
\multicolumn{16}{l}{Teachers (all trained on \imagenet)} \\
\invismidrule
\datarownumber & \dino \cite{caron2021emerging} &  &  & & 78.4 & & & \textbf{72.4} & & & 30.4 & & & 0.570 & \\
\datarownumber & \deit \cite{touvron2022deitiii} &  & & & 83.6 & & & 68.3 & & & 32.3 & & & 0.589 & \\
\datarownumber & \dbot \cite{liu2024dbot} &  &  & & \textbf{84.0} & & & \textbf{72.4} & & & 32.8 & & & 0.616  & \\
\datarownumber & \ibot \cite{zhou2022ibot} &  & & & 79.2 & & & 70.7 & & & \textbf{36.6} & & & \textbf{0.524} & \\

\invismidrule
\datarownumber & \multicolumn{3}{l}{{\em Max teacher performance on each task}} & & 84.0 & & & 72.4 & & & 36.6 & & & 0.524 & \\
\datarownumber & \multicolumn{3}{l}{{\em Min teacher performance on each task}} & & 78.4 & & & 68.3 & & & 30.4 & & & 0.616 & \\

\midrule
\multicolumn{16}{l}{Students distilled using real \imagenet data - \textbf{oracle}} \\
\invismidrule
\datarownumber & {UNIC} \cite{sariyildiz2024unic} &\multirow{5}{*}{\imagenet} & 1.28M & & 83.3 & & & 73.3 & & & 39.5 & & & 0.523 & \\
\cmidrule{4-16}
\datarownumber &  &  & 1M & & \gain{\bf 83.2} & & & \gain{\bf 73.1} & & & \gain{\bf 39.0}  & & & \gain{\bf 0.531}  & \\
\datarownumber &  &  & 100K & & \gain{\bf 81.7} & & & \gain{\bf 71.6} & & & \gain{\bf 37.8} & & & \gain{\bf 0.542}  & \\
\datarownumber &  &  & 10K & & {\bf 77.0}  & & & 66.5 & & & \gain{\bf 37.4} & & & \gain{\bf 0.545}  & \\
\datarownumber &  &  & 1K & & 72.6 & & & 65.3 & & & \gain{\bf 34.2} & & & \gain{\bf 0.566}  & \\

\midrule
\multicolumn{16}{l}{Students distilled using procedural synthetic data} \\
\invismidrule
\datarownumber &  & \multirow{4}{*}{Gaussian Noise} & 1M & & 22.6 & & & 30.9 & & & \,\,\,8.3 & & & 0.924 & \\
\datarownumber &  &  & 100K & & 22.7 & & & 31.3 & & & \,\,\,8.2 & & & 0.915  & \\
\datarownumber &  &  & 10K & & 21.6 & & & 30.4 & & & \,\,\,7.6 & & & 0.940  & \\
\datarownumber &  &  & 1K & & 21.1 & & &  29.4 & & &  \,\,\,7.2 & & &  0.965  & \\

\cmidrule{3-16}
\datarownumber &  & \multirow{4}{*}{Fractals} & 1M & & 32.7 & & & 37.3 & & & 10.9 & & & 0.932 & \\
\datarownumber &  &  & 100K & & 27.5 & & &  31.7 & & &  \,\,\,9.6 & & &  0.935  & \\
\datarownumber &  &  & 10K & & 27.1 & & &  31.7 & & &  \,\,\,9.4 & & &  0.925  & \\
\datarownumber &  &  & 1K & & 28.0 & & &  32.2 & & &  10.0 & & &  0.936  & \\

\cmidrule{3-16}
\datarownumber &  & \multirow{4}{*}{Dead Leaves} & 1M & & 66.9 & & & 65.5 & & & 29.1 & & & 0.632 & \\
\datarownumber &  &  & 100K & & 67.1 & & &  65.6 & & &  29.0 & & &  0.639  & \\
\datarownumber &  &  & 10K & & 66.6 & & &  65.1 & & &  29.1 & & &  0.655 & \\
\datarownumber &  &  & 1K & & 64.7 & & &  63.9 & & &  29.0 & & &  0.638 & \\

\midrule
\multicolumn{16}{l}{Students distilled using images optimized for the 4 teachers (\textbf{ours})} \\
\invismidrule
\datarownumber &  & \multirow{4}{3cm}{Improved Dead Leaves (\ours)} & 1M   & & 78.2 & \improv{17} & & \gain{70.6} & \improv{8} & & \gain{34.5} & \improv{18} & & \gain{0.573} & \improv{9} \\
\datarownumber &  &                                                 & 100K & & 78.0 & \improv{16} & & \gain{70.3} & \improv{7} & & \gain{34.5} & \improv{19} & & \gain{0.573} & \improv{10} \\
\datarownumber &  &                                                 & 10K  & & {\bf 77.0} & \improv{16} & & \gain{\bf 69.9} & \improv{7} & & \gain{34.5} & \improv{18} & & \gain{0.592} & \improv{10} \\
\datarownumber &  &                                                 & 1K   & & {\bf 74.1} & \improv{14} & & \gain{\bf 68.6} & \improv{7} & & \gain{33.9} & \improv{17} & & \gain{0.594} & \improv{7} \\
\bottomrule
\end{tabular}
}

\label{tab:all_results}
\end{table}

\paragraph{Evaluation.}
We evaluate all models on the same downstream tasks as UNIC~\cite{sariyildiz2024unic}.
This includes image classification on the \imagenet validation set, transfer learning, semantic segmentation, and depth estimation.
For transfer learning, we assess generalization on 15 image classification benchmarks: the 5 concept generalization levels of ImageNet-CoG~\cite{sariyildiz2021concept}, 8 fine-grained datasets (Aircraft~\cite{maji2013aircraft}, Cars196~\cite{krause2013cars}, DTD~\cite{cimpoi2014texture}, EuroSAT~\cite{helber2019eurosat}, Flowers~\cite{nilsback2008flowers}, Pets~\cite{parkhi2012cats}, Food101~\cite{bossard2014food101}, and SUN397~\cite{xiao2010sun}), and 2 long-tailed datasets (iNaturalist~\cite{van2018inaturalist} 2018 and 2019). We report Top-1 accuracy for all classification tasks. For dense prediction, we report mIoU on ADE20K \cite{zhou2019semantic} and RMSE on NYUd \cite{silberman2012indoor}, following the patch-level classification protocol of~\cite{oquab2024dinov2}.
For all these tasks, we use linear probing on frozen encoder outputs.
\subsection{From real images to data-free distillation}
\label{sec:exp_real_synth}

In this section, we first distill using real-image subsets of decreasing size, establishing an upper bound for the rest of our experiments.
We then examine different types of procedural synthetic data and assess them in the context of multi-teacher distillation.

\paragraph{Distilling using real images, our oracle.}
Before considering data-free distillation, we assess the impact of reducing the distillation set when it consists of real images.
To this end, we create subsets of \imagenet of varying sizes (1K, 10K, 100K, and 1M) and distill with each.

We report the individual teacher results in \texttt{rows 1-4} of \cref{tab:all_results}, with \texttt{rows 5-6} giving the maximum and minimum across teachers, which serve as reference for assessing the students. The students distilled on subsets follow in \texttt{rows 8-11}, where \texttt{row 7} is the reproduced UNIC model~\cite{sariyildiz2024unic}, \ie the student distilled on the full \imagenet dataset. 
We observe that reducing the distillation set from 1 million to 100K images barely affects results, showing that nearly 90\% of \imagenet can be discarded with negligible impact.
When further reducing to 10K images or fewer, the drop becomes far more drastic.
This part of the table provides an upper bound on the results achievable for a given budget, and serves as our oracle for the rest of the discussion.

\begin{figure}[t!]
    \centering
    \includegraphics[width=\columnwidth]{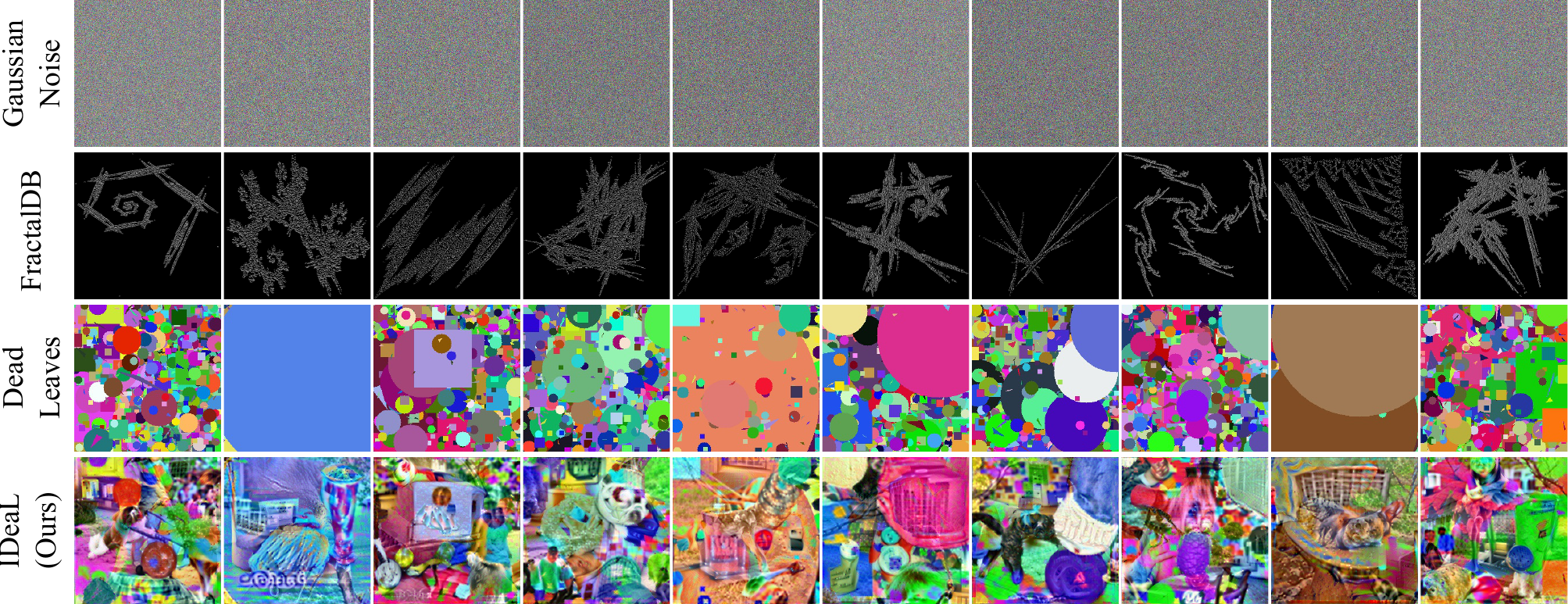}
    \caption{Samples from the synthetic datasets considered in our experiments. From top to bottom: Gaussian noise, \fractals, Dead Leaves, and our proposed Improved Dead Leaves (\ours). \ours optimizes Dead Leaves as described in \cref{sec:method}, providing significantly richer information to the student while remaining fully data-free.}
   \label{fig:synthetic_data}
\end{figure}

\paragraph{Distilling with procedural synthetic data.}
Next, we replace real images with procedural synthetic data during distillation.
Unlike synthetic data obtained with a generator trained on real images~\cite{sariyildiz2023fake}, procedural synthetic image generation~\cite{kataoka20pre-training,nakashima21can} or structured noise \cite{baradad21learning} processes construct samples directly from predefined rules without learning from any real image dataset. While these samples have similar statistics to real images such as power spectrum \cite{baradad21learning}, they do not contain any semantic content.
Additionally, no privileged information from the original training dataset is used, such as the class information or any other dataset specific statistics.
Inspired by \cite{baradad21learning}, we consider Gaussian noise, \fractals, and Dead Leaves as synthetic surrogate data for multi-teacher distillation (see examples in the first three rows of \cref{fig:synthetic_data}).


\begin{table}[t]
\setcounter{datarow}{0}
\centering
\caption{
Performance of different student models  distilled from the same four teachers, using varied flavors of Improved Dead Leaves as distillation data.
Each distillation run is performed with a different set of 10K samples.
Each set is obtained by optimizing with respect to a given set of teachers.
The column \texttt{Joint Optim.} indicates whether the samples are optimized with respect to the teachers jointly ($\checkmark$) or not.
\texttt{Row 9} combines 4 sets of 2.5K samples, each optimized for one teacher only,
whereas the 10K samples of \texttt{row 10} are obtained by optimizing with respect to the 4 teachers jointly.
The best performance per column is in \textbf{bold}.
}
\footnotesize
\setlength{\tabcolsep}{2.5pt}
\adjustbox{max width=\textwidth}{

\begin{tabular}{r l c c c c c c c}
\toprule

 & \multirow{2}{4cm}{Teachers used during sample optimization} & \multirow{2}{1cm}{Joint optim.} & \imagenet & Transfer & Seg. & Depth \\
 & & & Top-1($\uparrow$) & Top-1 ($\uparrow$) & mIoU ($\uparrow$) & RMSE ($\downarrow$) \\

\midrule
\multicolumn{7}{l}{Improved Dead Leaves optimized with 1 teacher} \\
\midrule
\datarownumber & \dino \cite{caron2021emerging}  & - & 74.4 & 69.2 & 33.0 & 0.584 \\
\datarownumber & \deit \cite{touvron2022deitiii} & - & 74.3 & 68.9 & 32.2 & 0.597  \\
\datarownumber & \ibot \cite{zhou2022ibot}       & - & 74.9 & 69.1 & 33.9 & 0.578  \\
\datarownumber & \dbot \cite{liu2024dbot}        & - & 74.9 & 69.3 & 32.5 & 0.599  \\
\midrule
\multicolumn{7}{l}{Improved Dead Leaves optimized with 2 teachers} \\
\midrule
\datarownumber & \dino, \dbot & $\checkmark$ & 75.9 & {\bf 70.0} & 34.0 & 0.605  \\
\datarownumber & \dino, \deit & $\checkmark$ & 75.5 & 69.3 & 33.1 & 0.589  \\
\datarownumber & \ibot, \dbot & $\checkmark$ & 76.0 & 69.6 & 34.2 & 0.575  \\
\datarownumber & \ibot, \deit & $\checkmark$ & 75.8 & 69.7 & 34.1 & {\bf 0.573}  \\
\midrule
\multicolumn{7}{l}{Improved Dead Leaves optimized with 4 teachers} \\
\midrule
\datarownumber & \dino, \deit, \ibot, \dbot  & -   & 75.1 & 69.4 & 33.9 & 0.583  \\
\datarownumber & \dino, \deit, \ibot, \dbot  & $\checkmark$ & {\bf 77.0} & 69.9 & {\bf 34.5} & 0.592  \\

\bottomrule
\end{tabular}
}
\label{tab:generation_ablation_affect_of_teachers}
\end{table}

We compare the performance of students distilled with these different synthetic sets, \texttt{rows 12-23} of \cref{tab:all_results}, to the students distilled with real data subsets, \texttt{rows 7-11}.
We observe that students distilled from Gaussian noise yield poor results, regardless of the size of the distillation set.
\fractals performs better, but still lags behind Dead Leaves.
Dead Leaves perform best among structured noise alternatives, and produce surprisingly strong transfer learning results.
For instance, 100K Dead Leaves achieves 65.6 in transfer, on par with the 65.3 obtained with a 1K subset of \imagenet (our oracle using real images).
Consequently, we use Dead Leaves as our starting point for the results presented in the next section.
\subsection{Tailoring Dead Leaves for multi-teacher distillation}
\label{sec:exp_trained_dead_leaves}

Dead Leaves yield strong results for a structured-noise-based distillation set; however, they are not yet competitive with real data.
In this section, we enhance Dead Leaves using our method (\cref{sec:method}) and report results in \cref{tab:all_results} (\texttt{rows 24-27}).
We make the following key observations.

\paragraph{Observation 1:}
\textit{Improved Dead Leaves (\ours) outperform the original Dead Leaves in all settings and for all tasks.}
On the \imagenet classification task, the corresponding student improves for instance by $14\%$ when using 1K samples (\texttt{row 23} vs. \texttt{row 27}), and by $17\%$ (\texttt{row 20} vs. \texttt{row 24}) when using 1M samples.
The largest improvements are observed for the semantic segmentation task, $18\%$ relative gain for 1M setting.
For transfer learning, the gains are smaller than for other tasks. To analyze this, we break down the transfer learning tasks into concept generalization, fine-grained, and long-tail classification. Across all three, \ours improves over Dead Leaves, by an overall 7\%.
The exception is long-tail tasks at larger subset sizes (100K and 1M), where \ours yields a 12\% gain, while smaller subsets follow the trend. We refer the reader to the supplementary material for the per-sub-task results.

\paragraph{Observation 2:}
\textit{Student models distilled on \ours outperform the minimum teacher performance for 3 tasks, even with as few as 1K samples.}
Comparing student and minimum teacher scores on each task reveals significant gains.
Given that students trained with \ours do not have access to any information about the nature of the teachers or the data they were trained on, we find these results impressive.
Even the students trained only with 1K \ours samples surpass the minimum teacher performance in transfer learning, semantic segmentation and depth estimation.
Specifically, in transfer learning, we observe a $+2.3$ increase in Top-1 accuracy in the 1M setting, closing the gap with the maximum teacher performance.
Moreover, in semantic segmentation, with a $+4.1$ gain in mIoU, the student outperforms all teachers except the top-performing one.
Only on the \imagenet classification task are student models trained on \ours samples slightly below the minimum teacher performance (\ie 78.2 \vs 78.4 when using 1M samples).
This is not surprising, as all four teachers were trained on that dataset, making it inherently challenging to close this gap using only structured noise. However, we emphasize that, given access to complementary teacher models and no additional information, one can still distill a single versatile student with only 1K \ours samples that is more balanced across tasks than any individual teacher.

\paragraph{Observation 3:}
\textit{In data scarce settings, students distilled with \ours perform on par with, or even better than students distilled on \imagenet subsets, in classification tasks.}
Here, we focus on extreme cases with access to only 1K and 10K images as distillation datasets.
Students distilled on \ours subsets outperform those distilled on \imagenet subsets by +3.3 and +3.4 Top-1 accuracy on the 1K and 10K sets, respectively, for transfer learning.
On \imagenet classification, the 1K \ours student outperforms the 1K \imagenet student by +1.5 accuracy.
Moreover, the 10K \ours students match the 10K \imagenet ones.

These results demonstrate that the optimized samples are highly informative, effectively encoding information from the teachers' representations, and enabling students to capture more of their knowledge.
However, this does not generalize to larger subsets (100K or 1M). Similar to the scaling trend noted in \cite{baradad21learning}, procedural synthetic data, including \ours, does not scale as well as \imagenet, particularly for dense tasks. We provide a scaling analysis in the supplementary material.
\subsection{Distilling to different architectures}
To ensure a fair evaluation of \ours, all experiments until here were conducted within the UNIC~\cite{sariyildiz2024unic} framework without modification, using a ViT-B/16 student and ViT-B/16 teachers. Since \ours samples are optimized leveraging information from the teachers only, they do not depend on the student architecture and can in principle be used to distill these teachers into any student.
To verify this we distill the same four ViT-B/16 teachers into a smaller ViT-S/16 student, using the same \ours samples used to produce the results reported in  \cref{tab:all_results}. In \cref{tab:vit_small_student}, we report the performances for our smallest (1K) and largest (1M) subset sizes. We find that our observations 1 and 3 also hold in this setup. \ours outperforms Dead Leaves across all tasks. Moreover, for observation 3, the gains at 1K samples are even more pronounced than those observed for ViT-B/16 in \cref{tab:all_results}: \ours outperforms Dead Leaves by 20\% on \imagenet classification and 27\% on semantic segmentation, while also surpassing \imagenet on all tasks. Consistent with the scaling limitations of synthetic data, at larger subset sizes (1M), a performance gap with \imagenet images remains. 
\begin{table}[t!]
\centering
\caption{
Different student architecture (ViT-S/16). Distillation of the four ViT-B/16 teachers into a ViT-S/16 student under the UNIC setup. Metrics as in \cref{tab:all_results}, for the 1K and 1M subset sizes; \imagenet is the real-data upper bound. Relative improvements of \ours over Dead Leaves are shown in green.
}
\begin{tabular}{@{}l r p{0.1cm}@{}c@{}c p{0.1cm}@{}c@{}c p{0.1cm}@{}c@{}c p{0.1cm}@{}c@{}c@{}}
\toprule
Distillation data & Size & \multicolumn{3}{c}{\imagenet} & \multicolumn{3}{c}{Transfer} & \multicolumn{3}{c}{Seg.} & \multicolumn{3}{c}{Depth} \\
& & \multicolumn{3}{c}{Top-1($\uparrow$)} & \multicolumn{3}{c}{Top-1($\uparrow$)} & \multicolumn{3}{c}{mIoU($\uparrow$)} & \multicolumn{3}{c}{RMSE($\downarrow$)} \\
\midrule
ImageNet (Oracle) & 1M & & {\bf 80.8} & & & {\bf 69.6} & & & {\bf 36.6} & & & {\bf 0.572} & \\
Dead Leaves & 1M & & 59.2 & & & 60.1 & & & 24.2 & & & 0.673 & \\
IDeaL (Ours) & 1M & & 72.6 & \improv{23} & & 66.3 & \improv{10} & & 30.6 & \improv{26} & & 0.634 & \improv{6} \\
\midrule
ImageNet (Oracle) & 1K & & 53.7 & & & 49.6 & & & 27.6 & & & 0.640 & \\
Dead Leaves & 1K & & 55.8 & & & 57.9 & & & 23.3 & & & 0.690 & \\
IDeaL (Ours) & 1K & & {\bf 66.8} & \improv{20} & & {\bf 63.4} & \improv{9} & & {\bf 29.6} & \improv{27} & & {\bf 0.632} & \improv{8} \\
\bottomrule
\end{tabular}
\label{tab:vit_small_student}
\end{table}

\begin{figure}[t]
\begin{minipage}[c]{0.52\textwidth}
\centering
\captionof{table}{
        Ablation study on our proposed loss components during optimizing Dead Leaves samples. 
        The first row corresponds to the original Dead Leaves whereas the other rows correspond to optimized ones for different loss combinations.
        We generate 10K samples for each configuration.
        $\LPD$ and/or $\LID$ are always used together with $\LTV$.
        \label{tab:ablation_loss}
    }
    \centering
\scriptsize
\setcounter{datarow}{0}
\setlength{\tabcolsep}{2pt}
\begin{tabular}{c c c cc cc}
\toprule
& \multirow{2}{0.45cm}{$\LID$}
& \multirow{2}{0.45cm}{$\LPD$}
& \imagenet & Transfer & Seg. & Depth \\
& &
& Top-1($\uparrow$) & Top-1($\uparrow$) & mIoU($\uparrow$) & RMSE($\downarrow$) \\

\midrule
\datarownumber  & - & - & 66.6 & 65.1 & 29.1 &  0.655 \\
\datarownumber  & $\checkmark$ & - & 73.4 & 68.2  & 32.4 &  0.663 \\
\datarownumber  & - & $\checkmark$ & 75.4 & 69.1 & 33.4 & 0.595 \\
\datarownumber  & $\checkmark$ & $\checkmark$ &  \textbf{77.0} & \textbf{69.9} & \textbf{34.5} & \textbf{0.592} \\

\bottomrule
\end{tabular}

\end{minipage}
\hfill
\begin{minipage}[c]{0.45\textwidth}
    \centering
    \includegraphics[width=\linewidth]{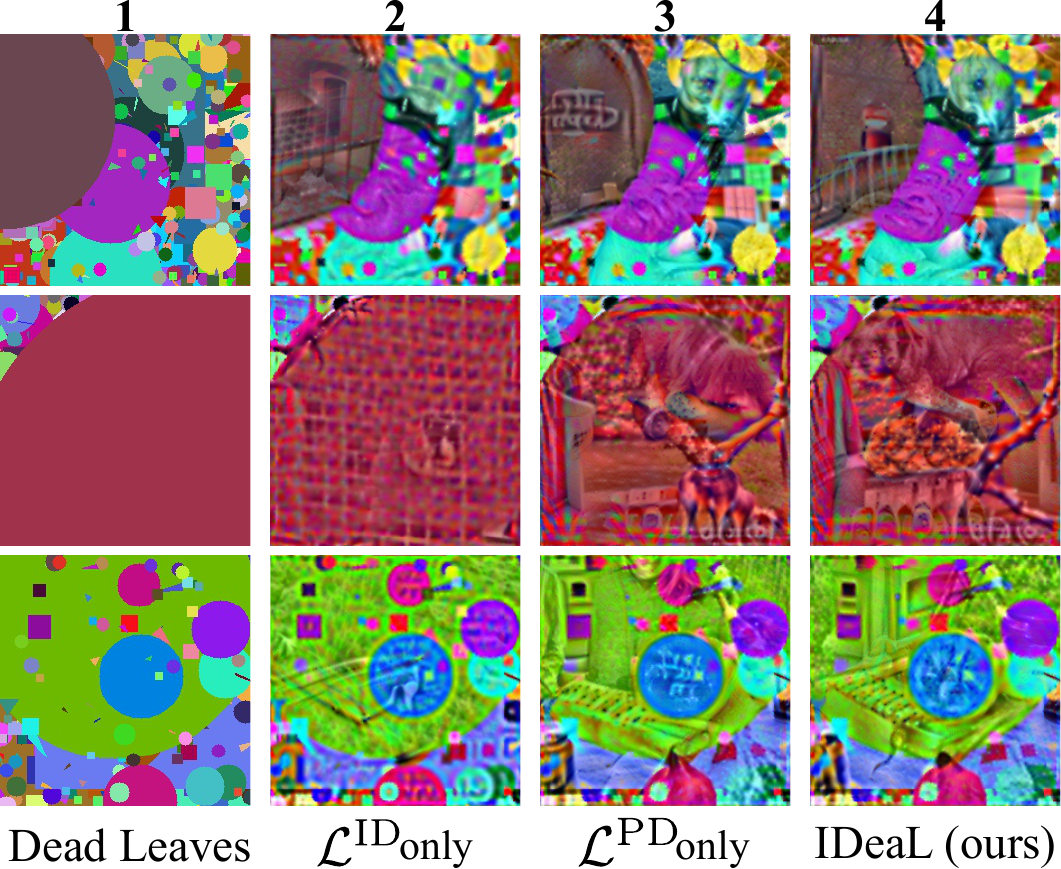}
    \captionof{figure}{
    Samples generated with different losses.
    Each loss contributes distinct structural properties.\label{fig:ablation_loss}}
\end{minipage}
\end{figure}

\paragraph{The exact set of teachers used during \ours generation matters.}
A question may arise: \textit{Would optimizing \ours for only a subset of the teachers lead to data that is equally effective for distillation?}
First, to investigate the effect of the set of teachers on the generated samples, in \cref{tab:generation_ablation_affect_of_teachers} we report results based on samples optimized using only one teacher (\texttt{rows 1--4}), two teachers (\texttt{rows 5--8}), and four teachers (\texttt{row 10}).
For all settings, combining more teachers boosts performance, and optimizing with all teachers produces the strongest results.
We also compare a different way of using all teachers.
In \texttt{row 9}, we combine an equal number of samples (2.5K) from each set of samples optimized for one teacher independently.
This set is then compared to the original \ours{} (\texttt{row 10}), which is jointly optimized with all teachers.
The clear difference between these two rows shows that our method effectively projects information from multiple teachers simultaneously.
As an additional remark, samples generated using \ibot, which performs best on dense tasks, lead to a student that also performs well on these tasks, compared to students trained on samples generated with other teachers.

\paragraph{Ablation on loss components.}
To analyze the impact of the components of our sample optimization loss \cref{eq:generation-loss}, we provide both qualitative (\cref{fig:ablation_loss}) and quantitative (\cref{tab:ablation_loss}) results.
Our reference point (\texttt{row 1}) is distilled using the original Dead Leaves dataset.
We compare this row to \ours optimized using only $\LID$ (\texttt{row 2}), only $\LPD$ (\texttt{row 3}), and using both (\texttt{row 4}), which corresponds to our standard \ours setting.
Note that the $\LTV$ term from \cref{eq:generation-loss} is always used for regularization.

Results show that $\LPD$ is the main source of performance gains, while $\LID$ provides a complementary boost.
Using $\LID$ alone is insufficient, particularly for dense tasks, and the best results are obtained when both losses are combined.
Additionally, the qualitative effect of each loss can be seen in \cref{fig:ablation_loss}, illustrating how each component contributes differently to the structure and properties of the generated images. Repeating this ablation with Gaussian noise instead of Dead Leaves as initialization yields the same conclusions; see the supplementary material for details.

\section{Conclusion}\label{sec:conclusion}
In this paper, we investigated the notion of data-free multi-teacher distillation. We first analyzed the effectiveness of 
structured noise in that context. Next, to close the performance gap between distilling with structured noise and with real images, we proposed a pixel-level optimization objective that improves structured noise in a way that effectively captures information from all teachers. We observed that these synthetic samples,
\ours, when used as distillation sets, significantly boost student performance across all tasks compared to a distillation set of non-optimized samples. 
Our optimization objectives are simple, require no teacher or dataset-specific information, and are truly data-free. As a result, they can potentially be applied to any other type of synthetic data, and extended to any kind of teachers or downstream tasks.


\bibliography{main}
\bibliographystyle{splncs04}

\renewcommand{\thefigure}{\Alph{figure}}
\renewcommand{\thetable}{\Alph{table}}

\clearpage
\appendix

\setcounter{figure}{0}    
\setcounter{table}{0}    
\begin{center}
    \Large Supplementary Material for \\
    \textbf{IDeaL: Data-Free Multi-Teacher Distillation \\
via Improved Dead Leaves}
\end{center}

\section{Additional details on multi-teacher distillation}
\label{sec:supp_dist_details}

This section extends Sec. 3 and Sec. 5.1 from the main paper by giving more details on the distillation framework that we use in all our experiments and adopted from UNIC \cite{sariyildiz2024unic}.

\paragraph{Distillation loss.} In Sec. 3 we defined the distillation loss $\Ldist = \sum_{i=1}^{\nt} \Lte{i}$. In UNIC, $\Lte{i}$ is the combination of a cosine similarity loss ($\Lcos$) and a smooth-$\ell_1$ loss ($\Lsl$), applied at both the \texttt{[CLS]} and patch levels:
\begin{equation}\label{eq:per-teacher-loss}
    \Lte{i}
    = \underbrace{\Lcos\!\left(\fclss,\, \fclst{i}\right)
    + \Lcos\!\left(\fpatchs,\, \fpatcht{i}\right)}_{\text{cosine similarity}}
    + \underbrace{\Lsl\!\left(\fclss,\, \fclst{i}\right)
    + \Lsl\!\left(\fpatchs,\, \fpatcht{i}\right)}_{\text{smooth-}\ell_1}.
\end{equation}
The cosine term encourages directional alignment between student and teacher features, while the smooth-$\ell_1$ term penalizes differences in magnitude.

\paragraph{Teachers.} As described in Sec. 5.1, we distill from 4 ViT-B/16 teachers, each pretrained on \imagenet \cite{deng2009imagenet}: two self-supervised models, (i) \dino \cite{caron2021emerging} and (ii) \ibot \cite{zhou2022ibot}, and two supervised models, (iii) \deit \cite{touvron2022deitiii} and (iv) \dbot \cite{liu2024dbot}, which is fine-tuned on \imagenet.
All teachers share the same encoder architecture: the patch size is 16 and the input image resolution is $224 \times 224$. 

\paragraph{Distillation setup for student.} The student is a ViT-B/16 model with the same architectural configuration as the teachers. During distillation, we enable the core UNIC components, including: (i) teacher-specific expendable projector heads, (ii) dedicated projectors for \texttt{[CLS]} and patch tokens, (iii) and the ladder-of-projectors architecture that connects intermediate student layers to the distillation objective. Note that projectors are omitted in Sec.~3 and \cref{eq:per-teacher-loss} for simplicity. We also apply feature standardization to teacher outputs and use teacher-dropping regularization. All projectors are discarded after distillation, and only the distilled encoder is used for downstream evaluation. We use the same distillation objective and apply teacher dropping with a probability $p=0.5$.
In \cite{sariyildiz2024unic}, the student is trained for 100 epochs unless stated otherwise, and for 200 epochs when teacher dropping regularization is enabled. In our experiments, we train all models for 100 epochs to keep the computational cost manageable. 
We use AdamW \cite{loshchilov2017adamW} with a base learning rate of $3e-4$ and cosine learning rate decay with linear warmup. The batch size is 128 per GPU and we use 4 GPUs. We apply the same data augmentation as in the original framework, including random resized crops, horizontal flips, color jitter, random grayscale conversion, Gaussian blur, and solarization. The augmentation pipeline is kept identical for all experiments for structured noise types, \ours and \imagenet to ensure fair comparisons.

\paragraph{Additional details on sample generation.}  In Sec 5.1, we already described the parameters that are used for our pixel optimization pipeline, and introduced the hyperparameters specific to this step. Please note that we do not report a detailed hyperparameter search as, in our observations, moderate changes to batch size, total iterations, or subset size only slightly affect downstream performance. Additionally, for the ablation, we also generated samples initialized from Gaussian noise without changing any hyperparameters.
\section{Detailed transfer learning results from Tab. 1}

In this section, we further present the experimental results for transfer learning performances of the students trained on different types of data. 
In Tab. 1, the main paper reports transfer learning evaluations as a single column averaged over 15 classification datasets. In \cref{tab:supp_all_results_gaussian_noise_included} we expand transfer results into three columns: concept generalization, fine-grained, and long-tail classification to provide the per-sub-task behavior already summarized in Sec. 5.3.
The breakdown shows that the smaller transfer learning gains over Dead Leaves are due to Dead Leaves already being a strong starting point on these tasks. Even without optimization, Dead Leaves is competitive with real data in data-scarce settings: at 1K samples it outperforms a 1K \imagenet subset by 4\% on long-tail classification, and the fine-grained gap is only 1\% (\texttt{row 11} vs. \texttt{row 23}). On top of this strong baseline, \ours still improves across all three sub-tasks, by an overall 7\% over Dead Leaves. The largest margin is on long-tail classification at the larger subset sizes (100K and 1M), where \ours reaches a 12\% gain, while the smaller subsets stay in line with the overall 7\% trend.
\section{Extended results with improved Gaussian noise} 
\label{sec:supp_extended_table} 
Recall that in Sec. 5.3, we made 3 key observations about
\ours: (1) \ours outperform the original Dead Leaves across all tasks, (2) students distilled on \ours outperform the minimum teacher performance across 3 tasks, even with the smallest subset size (1K), (3) in data-scarce settings, students distilled with \ours perform on par with, or better than students distilled on \imagenet subsets, in classification tasks. 

The effectiveness of our optimization method (which we used to build \ours from Dead Leaves) can also be observed on top of Gaussian noise.
\texttt{Rows 24-27} in \cref{tab:supp_all_results_gaussian_noise_included} present the distillation performance when Gaussian noise is used as the initial data source.
Specifically, focusing on Observation 1, 
our optimization objective significantly improves
the distillation set, and "improved Gaussian noise" also leads to stronger students. 
Note that the increase is larger than for Dead Leaves, since Gaussian noise is a much weaker starting point; 
 the performance gap between pure Gaussian noise and Dead Leaves is also significantly larger. 
When we compare the performance between the Gaussian Noise (\texttt{Rows 12-15}) and the improved Gaussian noise (\texttt{Rows 24-27}), the results further validate our claims: the optimization reduces the gap drastically by achieving more than a 2$\times$ improvement. 
\begin{table}[t!]
\centering
\caption{
\textbf{Extended version of Tab. 1 from the main paper}. 
Compared to the original table, transfer results are split into three columns (CoG, Fine-Grained, and Long-Tail).
The table shows the performance of teacher models and different students distilled from them, across different distillation data sources and sizes.
We report Top-1 accuracy on \imagenet, average Top-1 accuracy on transfer learning performance split into three columns for 5 concept generalization, 8 fine-grained, and 2 long-tail datasets,  mIoU on ADE20K for segmentation, and RMSE on NYUd for depth estimation.
In addition to Improved Dead Leaves (\ours), we also report Improved Gaussian Noise, i.e.\ our optimization method applied to Gaussian noise instead of Dead Leaves.
For students distilled on real \imagenet, Dead Leaves and \ours subsets, we report mean $\pm$ standard deviation over 3 random subsets. Best teacher performance and best student performance per distillation set size in \textbf{bold}.
Student performance better than the min teacher is \gaintext{highlighted in beige}.
}
\setcounter{datarow}{0}
\setlength{\tabcolsep}{2pt}
\adjustbox{max width=\textwidth}{
\footnotesize
\begin{tabular}{r l l r p{0.25cm}@{}c p{0.25cm}@{}c p{0.25cm}@{}c p{0.25cm}@{}c p{0.25cm}@{}c p{0.25cm}@{}c}
\toprule
& \multirow{2}{1.5cm}{Model} & \multicolumn{2}{c}{Distillation data} & \multicolumn{2}{c}{\imagenet} & \multicolumn{2}{c}{CoG} & \multicolumn{2}{c}{Fine-Grained} & \multicolumn{2}{c}{Long-Tail} & \multicolumn{2}{c}{Seg.} & \multicolumn{2}{c}{Depth} \\
& & Source & Size & \multicolumn{2}{c}{Top-1($\uparrow$)} & \multicolumn{2}{c}{Top-1($\uparrow$)} & \multicolumn{2}{c}{Top-1($\uparrow$)} & \multicolumn{2}{c}{Top-1($\uparrow$)} & \multicolumn{2}{c}{mIoU($\uparrow$)} & \multicolumn{2}{c}{RMSE($\downarrow$)} \\

\midrule
\multicolumn{16}{l}{Teachers (all trained on \imagenet)} \\
\invismidrule
\datarownumber & \dino \cite{caron2021emerging} &  &  & & 78.4 & & 65.3 & & \bf 81.6 & & \bf 53.1 & & 30.4 & & 0.570 \\
\datarownumber & \deit \cite{touvron2022deitiii} &  & & & 83.6 & & 64.0 & & 76.8 & & 44.8 & & 32.3 & & 0.589 \\
\datarownumber & \dbot \cite{liu2024dbot} &  &  & & \textbf{84.0} & & 65.8 & & 81.4 & & 52.4 & & 32.8 & & 0.616 \\
\datarownumber & \ibot \cite{zhou2022ibot} &  & & & 79.2 & & \bf 65.9 & & 78.8 & & 50.7 & & \textbf{36.6} & & \textbf{0.524} \\

\invismidrule
\datarownumber & \multicolumn{3}{l}{{\em Max teacher performance on each task}} & & 84.0 & & 65.9 & & 81.6 & & 53.1 & & 36.6 & & 0.524 \\
\datarownumber & \multicolumn{3}{l}{{\em Min teacher performance on each task}} & & 78.4 & & 64.0 & & 76.8 & & 44.8 & & 30.4 & & 0.616 \\

\midrule
\multicolumn{16}{l}{Students distilled using real \imagenet data - \textbf{oracle}} \\
\invismidrule
\datarownumber & {UNIC} \cite{sariyildiz2024unic} &\multirow{5}{*}{\imagenet} & 1.28M & & 83.3 & & 67.3 & & 81.9 & & 53.9 & & 39.5 & & 0.523 \\
\cmidrule{4-16}
\datarownumber &  &  & 1M & & \gain{\bf 83.2\tiny{$\pm$.09}} & & \gain{\bf 67.3\tiny{$\pm$.11}} & & \gain{\bf 81.6\tiny{$\pm$.24}} & & \gain{\bf 54.2\tiny{$\pm$.49}} & & \gain{\bf 39.0\tiny{$\pm$.13}} & & \gain{\bf 0.531\tiny{$\pm$.004}} \\
\datarownumber &  &  & 100K & & \gain{\bf 81.7\tiny{$\pm$.37}} & & \gain{\bf 66.2\tiny{$\pm$.51}} & & \gain{\bf 80.0\tiny{$\pm$.89}} & & \gain{\bf 51.0\tiny{$\pm$1.3}} & & \gain{\bf 37.8\tiny{$\pm$.86}} & & \gain{\bf 0.542\tiny{$\pm$.008}} \\
\datarownumber &  &  & 10K & & {\bf 77.0\tiny{$\pm$.20}} & & 62.7\tiny{$\pm$.26} & & 74.8\tiny{$\pm$.44} & & 43.0\tiny{$\pm$.95} & & \gain{\bf 37.4\tiny{$\pm$.04}} & & \gain{\bf 0.545\tiny{$\pm$.001}} \\
\datarownumber &  &  & 1K & & 72.6\tiny{$\pm$.13} & & 61.1\tiny{$\pm$.05} & & 73.9\tiny{$\pm$.06} & & 40.3\tiny{$\pm$.66} & & \gain{\bf 34.2\tiny{$\pm$.06}} & & \gain{\bf 0.566\tiny{$\pm$.001}} \\

\midrule
\multicolumn{16}{l}{Students distilled using procedural synthetic data} \\
\invismidrule
\datarownumber &  & \multirow{4}{*}{Gaussian Noise} & 1M & & 22.6 & & 27.3 & & 36.7 & & 9.6 & & \,\,\,8.3 & & 0.924 \\
\datarownumber &  &  & 100K & & 22.7 & & 27.2 & & 39.3 & & 9.4 & & \,\,\,8.2 & & 0.915 \\
\datarownumber &  &  & 10K & & 21.6 & & 26.3 & & 38.2 & & 9.4 & & \,\,\,7.6 & & 0.940 \\
\datarownumber &  &  & 1K & & 21.1 & & 25.6 & & 36.7 & & 9.7 & & \,\,\,7.2 & & 0.965 \\

\cmidrule{3-16}
\datarownumber &  & \multirow{4}{*}{Fractals} & 1M & & 32.7 & & 34.3 & & 44.8 & & 14.8 & & 10.9 & & 0.932 \\
\datarownumber &  &  & 100K & & 27.5 & & 28.6 & & 39.0 & & 10.1 & & \,\,\,9.6 & & 0.935 \\
\datarownumber &  &  & 10K & & 27.1 & & 29.0 & & 38.8 & & 10.1 & & \,\,\,9.4 & & 0.925 \\
\datarownumber &  &  & 1K & & 28.0 & & 30.0 & & 38.9 & & 10.9 & & 10.0 & & 0.936 \\

\cmidrule{3-16}
\datarownumber &  & \multirow{4}{*}{Dead Leaves} & 1M & & 66.9\tiny{$\pm$.43} & & 60.2\tiny{$\pm$.24} & & 74.3\tiny{$\pm$.37} & & 43.6\tiny{$\pm$0.84} & & 29.1\tiny{$\pm$.28} & & 0.632\tiny{$\pm$.007} \\
\datarownumber &  &  & 100K & & 67.1\tiny{$\pm$.26} & & 60.2\tiny{$\pm$.33} & & 74.1\tiny{$\pm$.40} & & 43.9\tiny{$\pm$1.2} & & 29.0\tiny{$\pm$.23} & & 0.639\tiny{$\pm$.007} \\
\datarownumber &  &  & 10K & & 66.6\tiny{$\pm$.25} & & 60.0\tiny{$\pm$.16} & & 73.7\tiny{$\pm$.15} & & 43.9\tiny{$\pm$.62} & & 29.1\tiny{$\pm$.17} & & 0.655\tiny{$\pm$.014} \\
\datarownumber &  &  & 1K & & 64.7\tiny{$\pm$.09} & & 58.9\tiny{$\pm$.03} & & 72.6\tiny{$\pm$.19} & & 41.9\tiny{$\pm$.20} & & 29.0\tiny{$\pm$.16} & & 0.638\tiny{$\pm$.001} \\

\midrule
\multicolumn{16}{l}{Students distilled using images optimized for the 4 teachers (\textbf{ours})} \\
\invismidrule
\datarownumber &  & \multirow{4}{3cm}{Improved Gaussian Noise} & 1M & & 77.3\tiny{$\pm$.35} & & 64.0\tiny{$\pm$.20} & & \gain{78.7\tiny{$\pm$.20}} & & \gain{48.2\tiny{$\pm$.84}} & & \gain{33.7\tiny{$\pm$.19}} & & \gain{0.577\tiny{$\pm$.008}} \\
\datarownumber &  & & 100K & & 77.2\tiny{$\pm$.10} & & 63.9\tiny{$\pm$.18} & & \gain{78.5\tiny{$\pm$.25}} & & \gain{48.1\tiny{$\pm$1.2}} & & \gain{33.6\tiny{$\pm$.34}} & & \gain{0.575\tiny{$\pm$.008}} \\
\datarownumber &  &  & 10K & & 76.2\tiny{$\pm$.15}  & & 63.3\tiny{$\pm$.21} & & \gain{78.0\tiny{$\pm$.13}} & & \gain{47.0\tiny{$\pm$.42}} & & \gain{33.3\tiny{$\pm$.18}} & & \gain{0.594\tiny{$\pm$.006}} \\
\datarownumber &  &  & 1K & & 73.3\tiny{$\pm$.17} & & 62.0\tiny{$\pm$.01} & & 76.3\tiny{$\pm$.04} & & 44.7\tiny{$\pm$.28} & & \gain{32.2\tiny{$\pm$.28}} & & \gain{0.600\tiny{$\pm$.009}} \\
\cmidrule{3-16}
\datarownumber &  & \multirow{4}{3cm}{Improved Dead Leaves (IDeaL)} & 1M & & 78.2\tiny{$\pm$.10} & & \gain{64.6\tiny{$\pm$.11}} & & \gain{79.8\tiny{$\pm$.12}} & & \gain{48.7\tiny{$\pm$.53}} & & \gain{34.5\tiny{$\pm$.23}} & & \gain{0.573\tiny{$\pm$.004}} \\
\datarownumber &  &  & 100K & & 78.0\tiny{$\pm$.20} & & \gain{64.5}\tiny{$\pm$.08} & & \gain{79.4}\tiny{$\pm$.18} & & \gain{48.2}\tiny{$\pm$.46} & & \gain{34.5}\tiny{$\pm$.20} & & \gain{0.573}\tiny{$\pm$.004} \\
\datarownumber &  &  & 10K & & {\bf 77.0\tiny{$\pm$.05}} & & \gain{\bf 64.1\tiny{$\pm$.09}} & & \gain{\bf 79.1\tiny{$\pm$.17}} & & \gain{\bf 47.3\tiny{$\pm$.55}} & & \gain{34.5\tiny{$\pm$.51}} & & \gain{0.592\tiny{$\pm$.003}} \\
\datarownumber &  &  & 1K & & {\bf 74.1\tiny{$\pm$.30}} & & \bf 62.8\tiny{$\pm$.20} & & \gain{\bf 78.1\tiny{$\pm$.21}} & &\gain{\bf 45.0\tiny{$\pm$.51}} & & \gain{33.9\tiny{$\pm$.18}} & & \gain{0.594}\tiny{$\pm$.004} \\
\bottomrule
\end{tabular}
}
\label{tab:supp_all_results_gaussian_noise_included}
\end{table}
\section{Ablation on losses with improved Gaussian noise}
\label{sec:supp_loss_gaussian}
\begin{figure}[t]
\begin{minipage}[c]{0.52\textwidth}
\centering
\captionof{table}{
        Extension of the ablation study on our proposed loss components during optimizing (Tab. 4) when applied to Gaussian noise. 
        The first row corresponds to the initial Gaussian noise whereas the other rows correspond to optimized ones with corresponding losses.
        We generate 10K samples for each configuration.
        $\LPD$ and/or $\LID$ are always used together with $\LTV$.
        \label{tab:supp_ablation_loss}
    }
    \centering
\scriptsize
\setcounter{datarow}{0}
\setlength{\tabcolsep}{2pt}
\begin{tabular}{c c c cc cc}
\toprule
& \multirow{2}{0.45cm}{$\LID$}
& \multirow{2}{0.45cm}{$\LPD$}
& \imagenet & Transfer & Seg. & Depth \\
& &
& Top-1($\uparrow$) & Top-1($\uparrow$) & mIoU($\uparrow$) & RMSE($\downarrow$) \\
\midrule
\datarownumber  & - & - & 21.6 & 30.4 & 7.6 & 0.940  \\
\datarownumber & $\checkmark$ & - & 68.4 & 64.5 & 29.1 & 0.636 \\
\datarownumber & - & $\checkmark$ & 75.2 & 68.6 & 32.0 & 0.578 \\
\datarownumber &  $\checkmark$ & $\checkmark$ & 76.2 & 69.0 & 33.3 & 0.594 \\

\bottomrule
\end{tabular}

\end{minipage}
\hfill
\begin{minipage}[c]{0.45\textwidth}
    \centering
    \includegraphics[width=\linewidth]{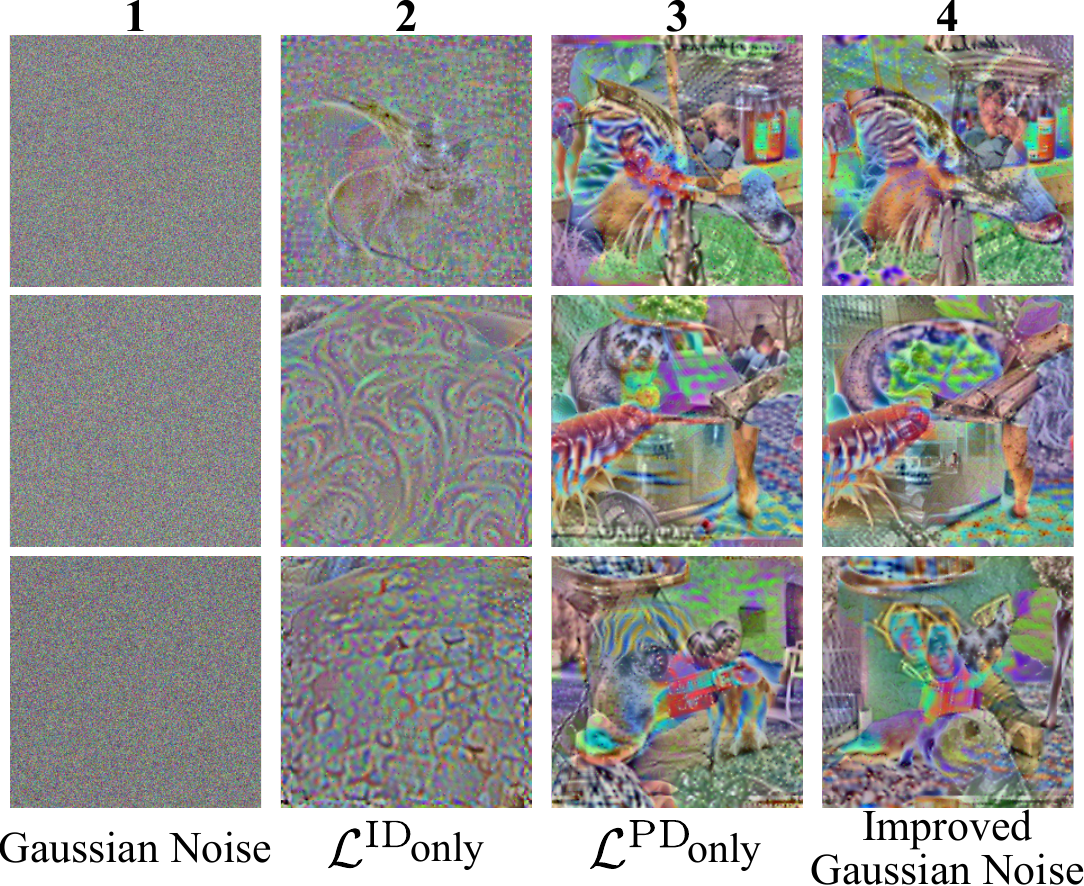}
    \captionof{figure}{
    Samples generated with different losses, starting from Gaussian noise.
    Each loss contributes distinct structural properties.\label{fig:supp_ablation_loss}}
\end{minipage}
\end{figure}

In Sec. 5.3, Tab. 4, and Fig. 4 of the main paper, we presented an ablation study on the effect of our proposed loss components, $\LID$ and $\LPD$. \cref{tab:supp_ablation_loss} and \cref{fig:supp_ablation_loss} extend this ablation to Gaussian Noise. Consistent with the findings in Tab. 4, the best results are obtained when both losses are used for optimization. Notably, even when using only $\LID$, which leverages information from the \texttt{[CLS]} tokens, we observe a substantial performance gain across all tasks, further highlighting the effectiveness of our optimization method.

\section{Scaling performance analysis}

\begin{figure}[t]

    \centering
    \includegraphics[width=\columnwidth]{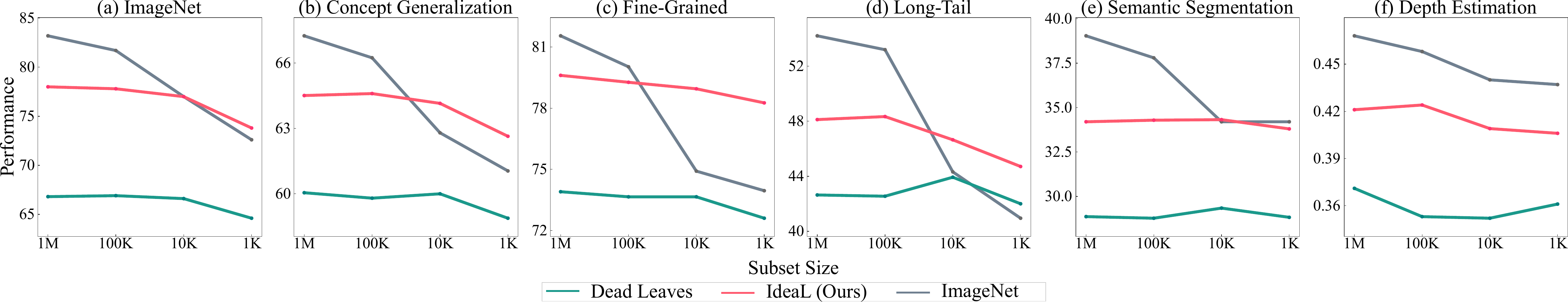}
    \centering
    \includegraphics[width=\columnwidth]{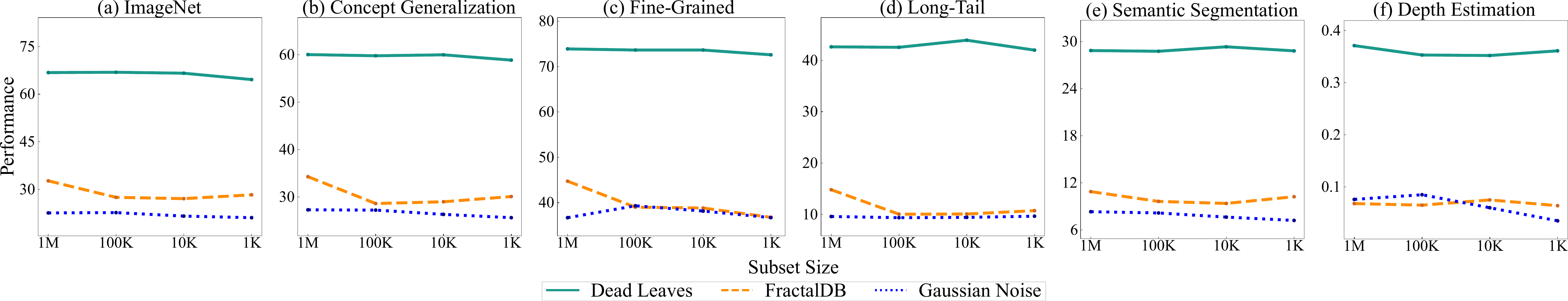}
    \caption{
    Scaling performance of the student using different training data. The bottom row demonstrates the performance with procedural synthetic data, while the top row compares performance when distilled on Dead Leaves, \ours, and \imagenet. Evaluation is performed on: (a) ImageNet classification, (b) concept generalization, (c) fine-grained, (d) long-tail, (e) semantic segmentation, and (f) depth estimation, across distillation set sizes 1K, 10K, 100K, and 1M. Exact values are provided in \cref{tab:supp_all_results_gaussian_noise_included}.}
    \label{fig:supp_line_plot_in1k_ideal_dl}

\end{figure}

This section discusses the scaling performance of \ours and of the procedural synthetic datasets when used for distillation. In \cref{fig:supp_line_plot_in1k_ideal_dl}, we provide two sets of plots, with different y-axes, to facilitate the comparisons. The bottom row shows scaling performance across procedural synthetic data, Dead Leaves, \fractals, and Gaussian noise, while the top row compares performance between Dead Leaves and \ours. 
We also include \imagenet as a reference.  
Similar to the scaling trend noted in \cite{baradad21learning}, we observe that procedural synthetic data does not scale well. 
When comparing \ours and Dead Leaves samples, \ours shows slightly better and more stable scaling,
especially on transfer learning tasks such as concept generalization and long-tail classification. However, when we analyze the performance of dense tasks (semantic segmentation and depth estimation),
\ours also struggles to scale up. 
This observation is more prominent when it is compared to \imagenet. 
Mitigating the poor scaling properties of synthetic data is still an open research question and is a promising direction for future work.

\section{Additional qualitative results}

\begin{figure}[t]
    \centering
    \includegraphics[width=\columnwidth]{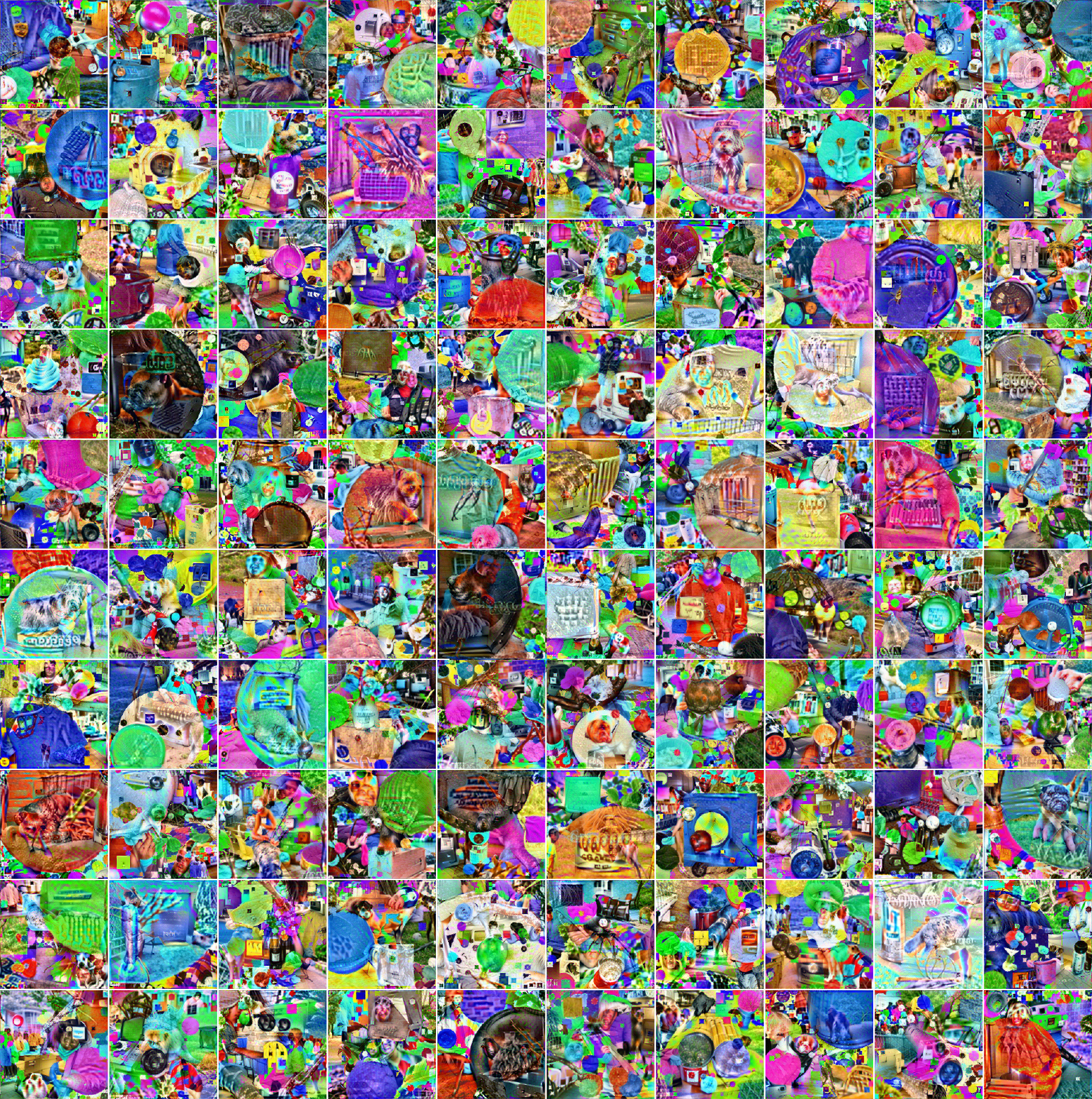}
   \caption{
    Qualitative examples of \ours samples initialized from Dead Leaves, obtained using our optimization method.}
    \label{fig:supp_qualitative_ideal}
\end{figure}

\begin{figure}[t]
    \centering
    \includegraphics[width=\columnwidth]{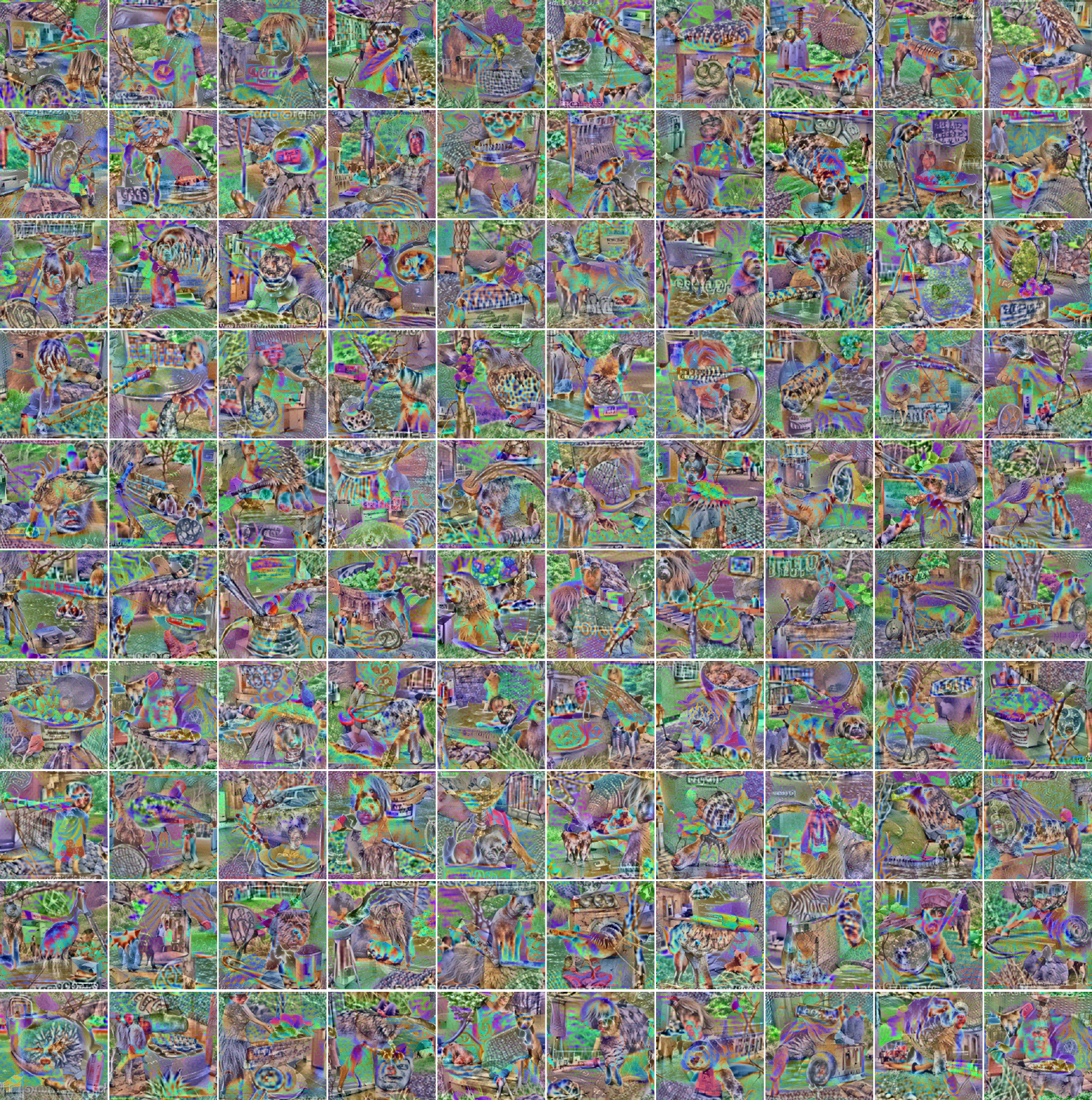}
   \caption{
    Qualitative examples of improved Gaussian noise samples initialized from Gaussian noise, obtained using our optimization method.}
    \label{fig:supp_qualitative_inoise}
\end{figure}

In this section, we provide qualitative examples, both from Improved Dead Leaves (\ours) and improved Gaussian noise, in \cref{fig:supp_qualitative_ideal} and \cref{fig:supp_qualitative_inoise}, respectively. Note that both sets of examples are generated using the exact same setup, only the initial images vary.

\label{sec:supp_exp_qualitative}

\end{document}